\documentclass{article}

\usepackage{PRIMEarxiv}

\usepackage[utf8]{inputenc} 
\usepackage[T1]{fontenc}    
\usepackage{hyperref}       
\usepackage{url}            
\usepackage{booktabs}       
\usepackage{amsfonts}       
\usepackage{nicefrac}       
\usepackage{microtype}      
\usepackage{lipsum}
\usepackage{fancyhdr}       
\usepackage{graphicx}       
\graphicspath{{media/}}     

\title{A Pipeline for Analysing Grant Applications
}

\author{
  Shuaiqun Pan, Sergio J. Rodríguez Méndez, Kerry Taylor \\
  Australian National University \\
  Canberra ACT 2601, AU\\
  \texttt{\{shuaiqun.pan, sergio.rodriguezmendez, kerry.taylor\}@anu.edu.au} \\
}

\begin{document}
\maketitle

\begin{abstract}
Data mining techniques can transform massive amounts of unstructured data into quantitative data that quickly reveal insights, trends, and patterns behind the original data. In this paper, a data mining model is applied to analyse the 2019 grant applications submitted to an Australian Government research funding agency to investigate whether grant schemes successfully identifies innovative project proposals, as intended. The grant applications are peer-reviewed research proposals that include specific ``innovation and creativity'' (IC) scores assigned by reviewers. In addition to predicting the IC score for each research proposal, we are particularly interested in understanding the vocabulary of innovative proposals. In order to solve this problem, various data mining models and feature encoding algorithms are studied and explored. As a result, we propose a model with the best performance, a Random Forest (RF) classifier over documents encoded with features denoting the presence or absence of unigrams. In specific, the unigram terms are encoded by a modified Term Frequency - Inverse Document Frequency (TF-IDF) algorithm, which only implements the IDF part of TF-IDF. Besides the proposed model, this paper also presents a rigorous experimental pipeline for analysing grant applications, and the experimental results prove its feasibility.
\end{abstract}

\keywords{Grant applications \and Random Forest classifier \and TF-IDF algorithm}

\section{Introduction}
\label{cha:intro}
In the 21st century, the importance of developing cutting-edge scientific research is self-evident for every country. Therefore, each country's government research funding agencies are willing to provide much scientific research funding to support essential and cutting-edge scientific research each year. Determining whether a scientific research project is worthy of funding is a significant and rigorous step for funding agencies. To obtain financial support, scientists and researchers always write research proposals to present their research plans and explain the significance of the project to the funding agencies \cite{denscombe2013role}. Usually, the government research funding agencies receive thousands of research proposals each year, which are reviewed only by expert panels. However, with the increase in the number of research proposals and the development of data mining techniques, funding agencies are increasingly using data mining models to assist in the manual review of research proposals. At the same time, it must be made clear that relying solely on data mining models to replace manual checks is not reliable.

Applying data mining models to a research proposal has several benefits. First, data mining models can briefly introduce the essential features of the research proposals to help human evaluators better screen the excellent research proposals, such as the influential features of the data mining models across all the research proposals. Second, an effective data mining model can help human evaluators understand the research proposals' strengths and weaknesses during the manual review process. Next, a high-quality data mining model can be applied to develop procedures and guidelines for human assessors to evaluate future research proposals to improve the quality of assessments. Fourth, for government or funding agencies, different funding projects should be established to improve the quality of various types of research. Data mining models can better understand how to ensure that human evaluators respond to these necessary qualities.

Based on the benefits and motivations mentioned above, we hope to apply a data mining model with an appropriate feature extraction technique to predict high IC-score research proposals based on the IC scores assigned by the expert reviewers. Meanwhile, the other primary goal of the project is to develop a predictive vocabulary for contemporaneous proposals and to understand how the model inferred research proposals with high IC scores from the data features. In addition, we focus on proposing an efficient feature extraction technique rather than the choice of classifiers, so we choose the very common Decision Tree (DT) and RF classifiers for experimental comparison.

The contributions of this paper are listed as follows:
\begin{itemize}
  \item A strict experimental pipeline for analysing grant applications is given, and the experimental results prove its feasibility.
  \item A model is proposed with a Random Forest classifier over documents encoded with features denoting the presence or absence of unigram terms. The unigram terms are encoded by a modified Term Frequency - Inverse Document Frequency (TF-IDF) algorithm, which only implements the IDF part of TF-IDF.
  \item The proposed model for predicting high IC-score research proposals can achieve an accuracy of 84.17\% across all types of grant applications.
\end{itemize}

This paper is divided into six sections. In the first section, the project's motivation and problem statement are briefly introduced. In section~\ref{cha:related}, the background and related work of this project are introduced. The methodology section mainly describes the pipeline we apply for this research project. Section~\ref{cha:experim} brings the overall design of the project. Then, the experimental settings and implementations with the hardware platforms are introduced in this section. The fifth section gives the experimental results of this project and carries on the further analysis. Conclusions and future work are described in section~\ref{cha:conclusion}.

\section{Related Work}
\label{cha:related}

\subsection{Computer science in evaluating grant applications}

Oztaysi et al. \cite{oztaysi2017evaluation} proposed a multi-criteria approach to evaluate research proposals based on interval-valued intuitionistic fuzzy sets. In this method, a fuzzy preference relation matrix was used to determine the relative importance of criteria. The Preference Selection Index (PSI) was another interesting method to evaluate research grant applications \cite{sundari2019decision}. One advantage of applying the PSI method was that the researcher did not need to determine the weight criteria. Another similar and recently related work was the research paper classification system built based on the TF-IDF and LDA schemes \cite{kim2019research}. This system used a Latent Dirichlet allocation (LDA) scheme to extract representative keywords from the abstract of each paper \cite{yau2014clustering}. The K-means clustering algorithm \cite{wagstaff2001constrained} was applied to group papers with similar topics based on the TF-IDF vector encoding of each paper. The results showed that the LDA with 30 keywords using TF-IDF obtained the best F-score compared with the LDA with fewer keywords.

\subsection{Term vectors and statistical measures in text representation}

TF-IDF is commonly used in data mining and information retrieval. TF indicates the frequency of a word in a document or a collection of documents. When calculating TF, all the words from documents are treated as equally important. However, in practice, people only pay attention to a certain of words. For example, ``this", ``are", and ``it" usually do not represent important in most cases. Then, the IDF is implemented to adjust the term weights in documents which can increase the weights of those rare but important words and weigh down those frequent words but less important. In 2016, Guo and Yang \cite{guo2016research} analysed the shortcomings of the TF-IDF algorithm. Then, an intra-class dispersion algorithm based on TF-IDF was proposed. Chen et al. \cite{chen2016turning} proposed a new term weighting technique called Term Frequency \& Inverse Gravity Moment (TF-IGM), which was mainly used to measure the class discrimination of a term. The experimental results showed that the TF-IGM performed better than the traditional TF-IDF in three standard corpora. Das and Chakraborty \cite{das2018improved} proposed a text sentiment classification technique based on the TF-IDF algorithm and Next Word Negation (NWN). In addition, this study also compared the binary bag of words, TF-IDF, and TF-IDF with NWN algorithms. Fan and Qin \cite{fan2018research} proposed another improved TF-IDF algorithm, TF-IDCRF, which focused on the relationship between classes in the classification model. In 2019, an improved TF-IDF algorithm based on classification discrimination strength was proposed for text classification \cite{zhang2019improved}.

\subsection{Data mining models in text classification}

In the field of data mining, the DT classifier is widely welcome for its advantage of showing how models make decisions according to the data features \cite{safavian1991survey}. RF classifier is another popular data mining model. The term forest can be interpreted to mean that each classifier in the ensemble is a DT classifier, while all combinations of classifiers are a forest \cite{han2011data}. In the RF classifier, each decision tree also selects the optimal attribute based on the Attribute Selection Measures (ASM). At the same time, each decision tree depends independently on a random sample. The RF classifier votes on each tree in specific classification problems and selects the most popular category as the final result.

In 2016, a news classification method was proposed based on the TF-IDF algorithm and Support Vector Machine (SVM) \cite{dadgar2016novel}. Based on a different number of n-grams and various data sets, five data mining classifiers were built and compared \cite{pranckevivcius2017comparison}. The results can guide researchers to select an appropriate data mining model according to the size of the data set. Four different data mining models were implemented with five different ensemble methods, and the experimental results showed that the RF classifier with the Bagging ensemble method achieved the best performance \cite{onan2016ensemble}. Wongso et al. \cite{wongso2017news} applied TF-IDF and SVD algorithms \cite{abdi2007singular} to the feature selection step and compare the two algorithms. At the same time, the Multivariate Bernoulli Naive Bayes \cite{kim2006some}, and SVM were compared in this study. Finally, with the combination of TF-IDF and Multivariate Bernoulli Naive Bayes, news articles in the Indonesian Language corpus were classified, and the best result was obtained \cite{wongso2017news}.

\section{Methodology}
\label{cha:method}

\begin{figure}[h]
\centering
\includegraphics[width=\textwidth]{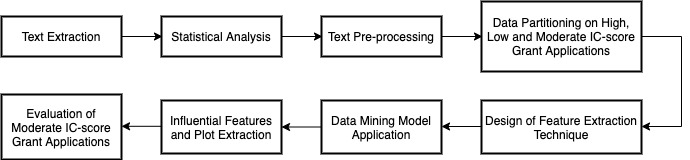}
\caption{The pipeline on analysing grant applications.}
\label{fig:1}
\end{figure}

This section details the workflow of our proposed pipeline and the data mining model. Fig.~\ref{fig:1} shows the pipeline of analysing grant applications.

\subsection{Data set}
\label{sec:dataset}
The data set used to analyse the grant applications is the 2019 grant applications submitted to an Australian Government research funding agency. 3,805 research proposals are given in this data set with peer-reviewed IC scores (1 - 7). In addition, the entire data set contains different types of grant applications, such as Synergy Grants, Standard Project Grants, and Ideas Grants. Besides the IC score, reviewers also score several other assessment scores, such as ``Feasibility Score'' or ``Significance Score.''

Since all research proposals are saved in PDF format, extracting the text from PDF files is necessary. A Metadata Extractor \& Loader (MEL) \cite{MEL} tool is applied to extract text from PDF research proposals and save it in a JSON file with metadata sets and content. By default, all JSON files are stored in CouchDB database \cite{couchdbapache} based on the proposal index.

Before designing the whole pipeline, a statistical analysis is required based on the IC scores of research proposals. At the same time, some fundamental values are also the basis of designing the entire pipeline, such as the median IC score and mode IC score. Table~\ref{table:3} shows a statistic of IC scores, showing that 3,693 research proposals have valid IC scores. In addition, 99 research proposals do not contain an IC score, 13 of which have an IC score below 1.0, and these research proposals therefore not be used in this project. Table~\ref{table:3} also shows the median IC score, 5.0, the most frequent IC score.

\begin{table}[]
\renewcommand\arraystretch{1.1}
\caption{Statistical summary of the IC scores of grant applications.}
\centering
\begin{tabular*}{0.4\textwidth}{@{}@{\extracolsep{\fill}}cc@{}}
\hline
Statistic & Value \\ 
\midrule
IC score counting & 3,693 \\
Standard deviation & 0.65 \\
Median & 5.00 \\
Mode & 5.00 \\
Mean & 4.92 \\
Min & 1.75 \\
25\% & 4.50 \\
50\% & 5.00 \\
75\% & 5.36 \\
Max & 6.90 \\
\hline
\end{tabular*}
\label{table:3}
\end{table}

\subsection{Text pre-processing for grant applications}
\label{sec:datapreprocessing}
HaCohen-Kerner et al. \cite{hacohen2020influence} proved that text pre-processing techniques could make the model achieve better performance than without the text pre-processing step. After all the text is extracted, all characters, whether uppercase or lowercase, are converted to lowercase. Then, the numbers are also removed because the numbers in the research proposals are not relevant for future analysis. Thirdly, removing punctuations and tokenizing by whitespace are also adopted, which make the text into small pieces called tokens.

In addition, the deletion of stop words is one of the most crucial text pre-processing techniques. Fani et al. \cite{fani2018stopword} have shown that deleting stop words can improve the performance of data mining tasks. Therefore, we create a list of custom stop words according to the IDF formula and delete the IDF value of the term from the text lower than 1.0. The reason for choosing 1.0 is that after implementation some preliminary experiments, we confirm that the feature words considered as important by DT and RF classifiers will less than 1000 words. Meanwhile, the words whose IDF value are less than 1.0 only account for 0.2\% of the total words, and they are all common words such as ``next'', ``shift'' and ``other''. We believe that these words appear too frequently and have no influence on the experimental results. Finally, text stemming is the last technique we apply in the text pre-processing step. Text stemming is a technique for reducing each word to its root format \cite{singh2016text}. It helps to reduce the vocabulary and surface syntax to get closer to the meaning of each term, and the Porter Stemming algorithm \cite{willett2006porter} is implemented in this step.

\subsection{Selection of high, low and moderate IC-score research proposals}
\label{sec:rightselection}
Assuming that the classification model is trained on research proposals with significantly high and low IC scores, better results will be obtained if indeterminate research proposals are avoided. Therefore, we conduct several experiments to determine the cut-off point to distinguish research proposals with high and low IC scores from moderate IC scores.

We decide to use the TF-IDF algorithm and RF classifier in the selection experiment. The training data range can be selected by training several cut-off points, such as 15, 20, 25, 30, 35, 40. Table \ref{table:1} shows a list of experiments for high, low, and moderate IC-score research proposals selection. For example, in the first experiment, all research proposals are ranked according to the IC score from 1 to 7, with 0$\sim$15\% as low IC-score research proposals and 85\%$\sim$100\% as high IC score research proposals. Then the selected research proposals are divided into 85\% training set and 15\% test set. In addition to the above research proposals, another 15\% of research proposals are randomly selected based on the original distribution in the moderate range. Then, the final test data combines the previous test data set and the selected 15\% moderate IC-score research proposals. It can be found that the actual size of the moderate IC-score research proposals and the final test data varies with the cut-off options. In the following experiments, each experiment is designed and conducted the same as the first experiment. If the number of intermediate-level research proposals is short, e.g., the No.5 and No.6 experiments, all moderate IC-score research proposals are considered.

\begin{table}[]
\renewcommand\arraystretch{1.1}
\caption{Experimental design of high, low and moderate IC-score research proposals selection.}
\centering
\begin{tabular*}{\textwidth}{@{}@{\extracolsep{\fill}}cccc@{}}
\hline
No. & Low IC score & High IC score & Moderate IC score \\
\midrule
1 & score\textless 4.25 (0$\sim$15\%) & score\textgreater 5.63 (85\%$\sim$100\%) & $5.63\ge $score$ \ge4.25$ (15\%$\sim$85\%) \\ 
2 & score\textless 4.36 (0$\sim$20\%) & score\textgreater 5.5 (80\%$\sim$100\%) & $5.5\ge $score$ \ge4.36$ (20\%$\sim$80\%) \\
3 & score\textless 4.5 (0$\sim$25\%) & score\textgreater 5.375 (75\%$\sim$100\%) & $5.375\ge $score$ \ge4.5$ (25\%$\sim$75\%)\\
4 & score\textless 4.57 (0$\sim$30\%) & score\textgreater 5.25 (70\%$\sim$100\%) & $5.25\ge $score$ \ge4.57$ (30\%$\sim$70\%)\\
5 & score\textless 4.75 (0$\sim$35\%) & score\textgreater 5.2 (65\%$\sim$100\%) & $5.2\ge $score$ \ge4.75$ (35\%$\sim$65\%)\\
6 & score\textless 4.75 (0$\sim$40\%) & score\textgreater 5.08 (60\%$\sim$100\%) & $5.08\ge $score$ \ge4.75$ (40\%$\sim$60\%)\\ 
\hline
\end{tabular*}
\label{table:1}
\end{table}

Some details require further explanation. Firstly, 800 research proposals (400 with low IC scores and 400 with high IC scores) are randomly selected to implement the RF classifier in each experiment. One reason is that it can reduce the training scale by drawing sampling from the training data set, thus reducing the model's time. The most important reason is that each research proposal has many unique terms, then there will be more than 110,000 unique terms in the corpus for 800 research proposals, which may lead to insufficient memory during model training.

Secondly, in the final test data set, the moderate IC-score research proposals are labelled based on the median IC score of 5.0. Proposals with an IC score greater than 5.0 are labelled as high IC-score proposals in the final test data set. If the research proposals have an IC score of less than 5.0, these proposals are labelled as low IC-score research proposals. Thirdly, it needs to be explained that research proposals with borderline IC scores are not selected when selecting research proposals with high or low IC scores. For example, the low IC score for experiment No.1 is set at 15\%, which is 4.25. Then, the proposals with an IC score of 4.25 will not be considered the low IC-score research proposals. Fourthly, the low IC score boundary of 4.75 is selected in both experiments 5 and 6. The reason is that the IC score of 4.75 is common, and more than 5\% (35\% $\sim$ 40\%) research proposals have this IC score.

\begin{table}[h!]
\caption{The selection experimental results based on RF classifier, TF-IDF algorithm and unigram features.}
\renewcommand\arraystretch{1.1}
\centering
\begin{tabular*}{0.3\textwidth}{@{}@{\extracolsep{\fill}}cc@{}}
\hline
No. & Test accuracy (\%) \\
\midrule
1 & 66.91 \\ 
2 & 65.88 \\
3 & 62.17 \\
4 & 56.84 \\
5 & 65  \\
6 & Less than 50 \\
\hline
\end{tabular*}
\label{table:17}
\end{table}

According to the results from Table~\ref{table:17}, the best performance can achieve 66.91\% based on an RF classifier, TF-IDF algorithm with 15\% and 85\% cut-off option. Then, the 15\%, 85\% option is selected for future experiments on different data mining models with feature extraction techniques.

\subsection{Design and apply the feature extraction technique}
\label{sec:newfeature}

We propose a modified TF-IDF algorithm, which only implements the IDF part of TF-IDF as the feature extraction technique. In specific, if the term exists at least once in the documents, specify the IDF value for this term directly. In addition, if a term does not exist in the documents, then the term is assigned a value of 0. The design of this modified feature extraction algorithm follows the idea that rare terms can define innovativeness.

The experiment also considers the n-grams \cite{tripathy2016classification}. Unigram is the most common choice for text classification tasks, but bigram and trigram may better represent scientific terms, where bigram is two consecutive words in a sentence, and trigram is three consecutive words in a sentence. At the same time, when collecting proposals, we also consider deleting the words that only exist once or twice, because very rare terms tend not to be predictive. In addition, the bigram mentioned in this paper denotes a combination of the unigrams and bigrams. The trigram denotes a combination of the unigrams, bigrams, and trigrams.

\subsection{Apply data mining models with grant applications}
\label{sec:datamining}
This paper uses DT and RF classifiers for text classification because we would like to find out the most influential terms and understood how the data mining model predicts high and low IC-score research proposals. The DT and RF classifiers are convenient to present this valuable information. Based on the experimental result of the high and low IC-score research proposals selection, all experiments are conducted with the low IC-score research proposals (IC score 0$\sim$15\%) and the high IC-score research proposals (IC score 85\%$\sim$100\%). In the comparison study of feature extraction techniques, 400 research proposals for each low and high IC score are randomly selected for model training, and the training data is 85\%, and the test data is 15\%. In order to analyse the proposed model in the end, the 100 most influential terms from the collections of research proposals are extracted by the function from scikit-learn library \cite{scikit-learn}, which bring us an intuitive understanding of how much each term contributes to reducing the weighted impurities.

\subsection{Analyse moderate IC-score grant applications}
\label{sec:analysemining}

We also conduct several experiments to analyse moderate IC-score research proposals based on the proposed model. The purpose of this series of experiments is to determine whether there is a relation between proposals with moderate IC scores and that of high and low IC scores. Since the proposed model is trained based on the low IC-score proposals of 0$\sim$15\% and high IC-score proposals of 85$\sim$100\%, the range of research proposals with moderate IC score is 15\%$\sim$85\%. Based on the median IC score, the selection range of testing moderate IC score by testing several cut-off options, such as 20, 25, 30, 35, 40, 45, and 50. Table \ref{table:2} shows a list of experiments used to analyse the research proposals of moderate IC score. Considering the symmetric distribution of the IC scores, new research proposals with low and high IC scores are selected in each experiment, and performance analysis is conducted based on the proposed training model. In addition to the experiments in Table~\ref{table:2},  another experiment is designed to check the median IC-score research proposals (IC score = 5.0) to predict the proportion of high or low IC-score research proposals rather than calculate the test accuracy.

\begin{table}[]
\caption{Experimental design for analyzing moderate IC-score research proposals.}
\centering
\begin{tabular*}{0.7\textwidth}{@{}@{\extracolsep{\fill}}ccc@{}}
\hline
No. & New low IC-score range & New high IC-score range \\ 
\midrule
1 & 15\%$\sim$20\% & 80\%$\sim$85\% \\
2 & 15\%$\sim$25\% & 75\%$\sim$85\% \\
3 & 15\%$\sim$30\% & 70\%$\sim$85\% \\
4 & 15\%$\sim$35\% & 65\%$\sim$85\% \\
5 & 15\%$\sim$40\% & 60\%$\sim$85\% \\
6 & 15\%$\sim$45\% & 55\%$\sim$85\% \\
7 & 15\%$\sim$50\% & 50\%$\sim$85\% \\
\hline
\end{tabular*}
\label{table:2}
\end{table}

\section{Experimental Settings}
\label{cha:experim}

This section describes all experimental settings for this paper. Initially, MEL \cite{MEL} is implemented through a set of Python-based methods to extract metadata for all supported file types. To extract metadata from the PDF version of a file, the Tesseract-OCR method \cite{textract} and pdftotext tool \cite{Xpdf} are applied. In the statistical analysis of grant applications, the Python language and Numpy library \cite{mckinney2012python} are used to calculate the median, mode, and other statistical measurements of IC score. In the experiments of selecting high, low, and moderate IC-score research proposals and implementing the data mining models, the scikit-learn library \cite{scikit-learn} is applied to implement the DT and RF classifiers. The python library gensim \cite{rehurek_lrec} is used to implement the TF-IDF algorithm and the newly proposed modified TF-IDF algorithm.

Hyper-parameter tuning is a significant step in applying data mining models, and the Bayesian Optimization tool \cite{snoek2012practical} is applied. The first step to implement Bayesian Optimization is to define the data mining model, such as the RF classifier and its parameters and corresponding bounds. In addition, we also need to implement the scoring method and the cross-validation setup. Secondly, the \emph{maximize} method is used to run the technique with \emph{n\_iter} and \emph{init\_points} parameters. The \emph{n\_iter} is defined for the number of steps to run the optimization function. The more steps, the easier it is to find the best accuracy value. The \emph{init\_points} is defined for random exploration on the parameter space, which helps to explore the diversity of the space. Finally, the parameter values for each accuracy are listed, highlighting the best combination of the parameter and the target value.

To find the hyper-parameters of the RF classifier, the range of each parameter is set as follows: max\_depth $=$ (5, 60), min\_samples\_split $=$ (10, 100), max\_features $=$ (0.1, 0.999), max\_samples\_leaf $=$ (10, 50) and n\_estimation $=$ (100, 400). For the DT classifier, the range settings for finding hyper-parameters are as follows: max\_depth $=$ (3, 10), min\_samples\_split $=$ (3, 10), max\_features $=$ (0.1, 0.999),and max\_samples\_leaf $=$ (3, 10). The max\_depth parameter indicates the maximum depth of the tree, and the max\_features denotes the number of features to consider when finding the best split \cite{scikit-learn}. The parameters min\_samples\_leaf, min\_samples\_split, and n\_estimators are defined as the minimum number of samples needed on a leaf node, the minimum number of samples needed to split an internal node, and the number of trees in the forest, respectively. All experiments related to RF classifier and DT classifier adopt the same setting of the hyper-parameter range. Meanwhile, the 10-fold cross-validation method is also applied in finding the hyper-parameters.

To evaluate the performance of the newly proposed modified TF-IDF algorithm and the TF-IDF algorithm with different data mining classifiers, the classification accuracy (Acc), F1 score are selected as the evaluation metrics. The hardware platform is MacBook Pro with Intel Core i7 2.9 GHz Quard-Core processor. The memory configuration is 16GB 2133 MHz LPDDR3.

\section{Experimental Result}
\label{cha:result}

\begin{table}[]
\caption{Experimental results with TF-IDF algorithm.}
\centering
\begin{tabular*}{\textwidth}{@{}@{\extracolsep{\fill}}cccc@{}}
\hline
No. & Model & Acc(\%) & F1 score(\%) \\
\midrule
1 & DT classifier + unigram & 76.67 & 71.43 \\
2 & DT classifier + unigram + remove words exist $\leq$ 1 time & 76.67 & 71.43 \\
3 & DT classifier + unigram + remove words exist $\leq$ 2 times & 79.17 & 76.47 \\
4 & DT classifier + bigram & 75.83 & 76.03 \\
5 & DT classifier + bigram + remove words exist $\leq$ 1 time & 73.33 & 68.63 \\
6 & DT classifier + bigram + remove words exist $\leq$ 2 times & 71.67 & 66.67 \\
7 & DT classifier + trigram & 76.67 & 71.43 \\
8 & DT classifier + trigram + remove words exist $\leq$ 1 time & 76.67 & 71.43 \\
9 & DT classifier + trigram + remove words exist $\leq$ 2 times & 76.67 & 71.43 \\
10 & RF classifier + unigram & 81.67 & 78.85 \\
11 & RF classifier + unigram + remove words exist $\leq$ 1 time & 80.83 & 77.67 \\
12 & RF classifier + unigram + remove words exist $\leq$ 2 times & 80 & 76.47 \\
13 & RF classifier + bigram & 80.83 & 80.34 \\
14 & RF classifier + bigram + remove words exist $\leq$ 1 time & 80.83 & 80.00 \\
15 & RF classifier + bigram + remove words exist $\leq$ 2 times & 79.17 & 77.88 \\
16 & RF classifier + trigram & 81.67 & 78.85 \\
17 & RF classifier + trigram + remove words exist $\leq$ 1 time & 80 & 76.47 \\
18 & RF classifier + trigram + remove words exist $\leq$ 2 times & 81.67 & 78.85 \\
\hline
\end{tabular*}
\label{table:6}
\end{table}

Table~\ref{table:6} shows the performance of the TF-IDF algorithm with DT and RF classifiers. It can be found that the RF classifier can consistently achieve better performance than the DT classifier under the different settings of the n-grams and deletion of rare terms.

\begin{table}[]
\caption{Experimental results with the newly proposed modified TF-IDF algorithm.}
\centering
\begin{tabular*}{\textwidth}{@{}@{\extracolsep{\fill}}cccc@{}}
\hline
No. & Model & Acc(\%) & F1 score(\%) \\ 
\midrule
1 & DT classifier + unigram & 77.5 & 74.29 \\
2 & DT classifier + unigram + remove words exist $\leq$ 1 time & 80.83 & 78.10 \\
3 & DT classifier + unigram + remove words exist $\leq$ 2 times & 77.5 & 75.23 \\
4 & DT classifier + bigram & 79.17 & 75.73 \\
5 & DT classifier + bigram + remove words exist $\leq$ 1 time & 79.17 & 77.48 \\
6 & DT classifier + bigram + remove words exist $\leq$ 2 times & 79.17 & 73.68 \\
7 & DT classifier + trigram & 76.67 & 69.57 \\
8 & DT classifier + trigram + remove words exist $\leq$ 1 time & 76.67 & 69.57 \\
9 & DT classifier + trigram + remove words exist $\leq$ 2 times & 74.17 & 73.50 \\
10 & RF classifier + unigram & 84.17 & 81.55 \\
11 & RF classifier + unigram + remove words exist $\leq$ 1 time & 84.17 & 81.55 \\
12 & RF classifier + unigram + remove words exist $\leq$ 2 times & 84.17 & 81.55 \\
13 & RF classifier + bigram & 84.17 & 81.55 \\
14 & RF classifier + bigram + remove words exist $\leq$ 1 time & 83.34 & 80.39 \\
15 & RF classifier + bigram + remove words exist $\leq$ 2 times & 84.17 & 81.55 \\
16 & RF classifier + trigram & 84.17 & 81.55 \\
17 & RF classifier + trigram + remove words exist $\leq$ 1 time & 84.17 & 81.55 \\
18 & RF classifier + trigram + remove words exist $\leq$ 2 times & 84.17 & 81.55 \\
\hline
\end{tabular*}
\label{table:7}
\end{table}

Table~\ref{table:7} shows the performance of the newly proposed modified TF-IDF algorithm with DT and RF classifiers. Based on the comparison of Table~\ref{table:6} and Table~\ref{table:7}, the best performance is achieved with 84.17\% accuracy by the RF classifier with the newly proposed modified TF-IDF algorithm except the No.14 model combination in Table~\ref{table:7}. The hyper-parameters are max\_depth = 22, max\_features = 0.9931, min\_samples\_leaf = 11, min\_samples\_split = 67 and n\_estimation = 102. To include all the terms from the corpus, we choose the RF classifier based on unigram and the modified TF-IDF algorithm as the final proposed model. Another reason why we do not choose the bigram and trigram combinations as the proposed model is the bigram and trigram terms are in fact not regarded as essential features by DT and RF classifiers. Features extracted from the proposed model shows that only 618 features are considered significant, based on tens of thousands of features in the research proposals.

Based on the comparison of the two tables, it can be found that the proposed modified TF-IDF algorithm is practical and effective despite two or three exceptions exist. At the same time, the experimental results prove that the core idea of defining the modified TF-IDF algorithm is meaningful and show the rare terms associated with innovativeness. It should also be noted that the newly proposed modified TF-IDF algorithm can be understood as a simple encoding technique, such as taking the value 0 or the IDF value of the term depending on whether the term exists in the research proposals. Based on the decision tree plots generated by the best performance model, it can be found that the modified TF-IDF algorithm does not affect the shape of the tree as seen in the tree graph, helping to understand whether the chosen split term is rare or common.

From the result of finding hyper-parameters, it can be found that the best performing model does not use all the features to apply with the data mining algorithms, such as the RF classifier only uses 99.31\% features. In addition, although we consider different n-grams, especially bigram and trigram, with removing scarce words, Table~\ref{table:6} and Table~\ref{table:7} could prove that it might help but not always. Moreover, based on the same feature extraction algorithm,  the classification accuracy of the RF classifier is always better than that of the DT classifier. Nevertheless, the results of the DT classifier are still crucial because the plot of DT classifier contains all the decisions.

\begin{figure}
\centering
\includegraphics[width=0.8\columnwidth]{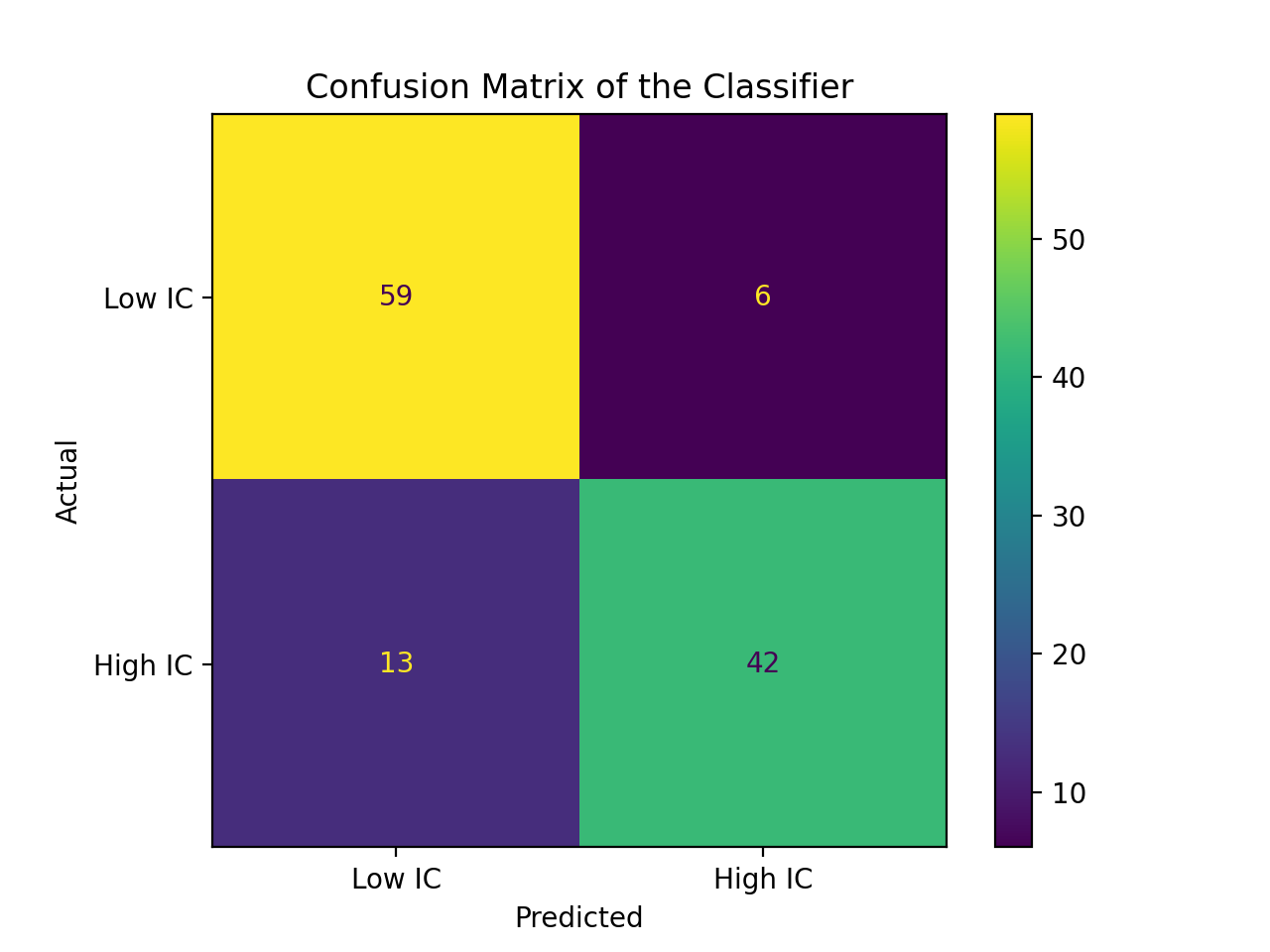}
\caption{Confusion matrix on ``unseen'' test set of the proposed model.}
\label{fig:17}
\end{figure}

Fig.~\ref{fig:17} shows the confusion matrix of the proposed model for the ``unseen'' test data. It shows 13 high IC-score research proposals are incorrectly predicted as low IC-score proposals. In addition, 6 research proposals with low IC scores are guessed wrongly which they are predicted as high IC-score proposals. The number 59 denotes that the proposed model correctly predicts 59 research proposals with low IC scores and 42 with high IC scores.

In addition to analysing the confusion matrix, we also extract the 100 most influential features from the proposed model, which gives an intuitive understanding of how much each feature contributes to reducing the weighted impurities. The top 100 features give us a better understanding of what is going on inside the black box. A measure of the feature importance is valuable for internal model development purposes by showing to what extent features contribute to test data. Although the classifier is only established for the 2019 grant applications and may not predict the high research proposals for future applications, these unique terms are still valuable and meaningful as a reference for evaluators.

Table~\ref{table:11} brings the performance of checking research proposals of moderate IC scores based on the proposed model. Based on the test accuracy, it can be concluded that there is a correlation between the moderate IC-score research proposals and high/low IC-score research proposals. Moreover, it is easy to find that the proposed model can better predict the research proposals close to the original training set settings (0$\sim$15\% for low IC score and 85\%$\sim$100\% for high IC score).

\begin{table}[]
\caption{Experimental results on analysing moderate IC-score research proposals.}
\centering
\begin{tabular*}{0.23\textwidth}{@{}@{\extracolsep{\fill}}cc@{}}
\hline
No. & Test Acc (\%) \\
\midrule
1 & 75.91 \\
2 & 73.43 \\
3 & 72.5 \\
4 & 65.63 \\
5 & 65.25 \\
6 & 64 \\
7 & 61.38 \\ 
\hline
\end{tabular*}
\label{table:11}
\end{table}

Based on the confusion matrix above and the experimental results of checking moderate IC-score research proposals, it can be found that the model is always more accurate in predicting research proposals with low IC scores than with high IC scores. Meanwhile, the research proposals with the median IC score of 5.0 are predicted to be about 37.2\% with high-IC score research proposals and about 62.8\% with low-IC score research proposals. Therefore, it can be concluded that research proposals with high IC scores use more diverse language than those with low IC-score. In addition to the experiments analysing all grant applications, we follow the same pipeline and establish a new model to evaluate Ideas Grant applications only, the one with innovation criteria. Applying the same method but with different hyper-parameters, the best performing model for analysing the Ideas Grants can reach an accuracy of 82.5\%. In every Ideas Grant application, there is a section called ``Innovation and Creativity statement.'' We also extract this part from each Ideas grant and analyse using the proposed pipeline. The experimental result shows that the proposed method can achieve 68.33\% accuracy on analysing ``Innovation and Creativity statement'' sections only from Ideas Grants. Although we guess the IC score is more relevant to the ``Innovation and Creativity statement'' compared with other sections, as evaluators may describe their innovation in this section, the experimental result does not support our guess.

\section{Conclusion}
\label{cha:conclusion}

In summary, a pipeline for analysing grant applications has been proposed with several crucial steps. The proposed data mining model is an RF classifier over documents encoded with features denoting the presence or absence of unigrams. Specifically, the unigram terms are encoded by a modified Term Frequency - Inverse Document Frequency(TF-IDF) algorithm, which only implements the IDF part of TF-IDF. As a result, the proposed model achieves an accuracy of 84.17\% based on all types of grant applications. In addition, we also build experiments for Ideas Grants only and ``Innovation and Creativity statement'' single section.

The future work can be carried out from different perspectives. Firstly, innovation should not be the only evaluation criterion. In order to better evaluate the entire grant application, we should consider other evaluation scores and establish a more comprehensive system that can predict a grant application based on multiple criteria. Secondly, in the future, this project can also apply some other effective data mining models, such as SVM, AdaBoost, and Xgboost. In addition, the pre-trained language models in the Natural Language Processing (NLP) field perform well in understanding text semantics, which can also be our next research focus. Thirdly, our proposed method cannot predict future grant applications because the current data set contains only key terms for 2019 and does not represent future grant applications. Therefore, it is an important research topic to consider building a long-term data mining model to predict future grant applications.


\bibliographystyle{unsrt}  
\bibliography{references}

\end{document}